\documentclass{article}
\usepackage{arxiv}

\usepackage{fancyhdr}
\usepackage[utf8]{inputenc} 
\usepackage[T1]{fontenc}    
\usepackage{hyperref}       
\usepackage{url}            
\usepackage{booktabs}       
\usepackage{amsfonts}       
\usepackage{nicefrac}       
\usepackage{microtype}      
\usepackage{algorithm}
\usepackage{subcaption}
\usepackage{algpseudocode}
\usepackage{lipsum}
\usepackage{graphicx}
\graphicspath{ {./images/} }

\title{Speak2Sign3D: A Multi-modal Pipeline for English Speech to American Sign Language Animation}

\author{
\textbf{Kazi Mahathir Rahman}$^{1}$, 
\normalfont Naveed Imtiaz Nafis$^{2}$, 
Md. Farhan Sadik$^{3}$,
Mohammad Al Rafi$^{4}$,\\
Mehedi Hasan Shahed$^{5}$ \\
Department of Computer Science, BRAC University, Dhaka, Bangladesh \\
\texttt{\{kazi.mahathir.rahman, naveed.imtiaz.nafis, md.farhan.sadik,} \\
\texttt{mohammad.al.rafi, mehedi.hasan.shahed\}@g.bracu.ac.bd}}

\begin{document}
\maketitle
\begin{abstract}
Helping deaf and hard-of-hearing people communicate more easily is the main goal of Automatic Sign Language Translation. Although most past research has focused on turning sign language into text, doing the reverse, turning spoken English into sign language animations, has been largely overlooked. That’s because it involves multiple steps, such as understanding speech, translating it into sign-friendly grammar, and generating natural human motion. In this work, we introduce a complete pipeline that converts English speech into smooth, realistic 3D sign language animations. Our system starts with Whisper to translate spoken English into text. Then, we use a MarianMT \cite{paper71} machine translation model to translate that text into American Sign Language (ASL) gloss, a simplified version of sign language that captures meaning without grammar. This model performs well, reaching BLEU \cite{paper73} scores of 0.7714 and 0.8923. To make the gloss translation more accurate, we also use word embeddings such as Word2Vec \cite{paper55} and FastText \cite{paper57} to understand word meanings. Finally, we animate the translated gloss using a 3D keypoint-based motion system trained on Sign3D-WLASL, a dataset we created by extracting body, hand, and face keypoints from real ASL videos in the WLASL \cite{paper42} dataset. To support the gloss translation stage, we also built a new dataset called BookGlossCorpus-CG, which turns everyday English sentences from the BookCorpus  \cite{paper77} dataset into ASL gloss using grammar rules. Our system stitches everything together by smoothly interpolating between signs to create natural, continuous animations. Unlike previous works like How2Sign \cite{paper67} and Phoenix-2014T \cite{paper78} that focus on recognition or use only one type of data, our pipeline brings together audio, text, and motion in a single framework that goes all the way from spoken English to lifelike 3D sign language animation.
\end{abstract}


\section{Introduction}
Imagine standing in a room full of people and hearing sounds but you do not know what those sounds are. This is the story of millions of deaf and hard-of-hearing people. However, being proficient in sign language, American Sign Language(ASL) is rich. Now imagine a technology that can fill this gap, using speech-to-hand translation to convert spoken words into live 3D hand signs in Sign Language. We aim to unfold the promising cross-section between deep learning and natural language processing (NLP) and how this makes the above vision possible to convert speech into 3D ASL and enhance access to the Deaf and Hard of Hearing (DHH).

\vspace{0.1cm}
ASL is a complicated visual-spatial language that uses the hand(s), facial expressions, and posture of the body for establishing contextual meanings in sign-stylized and lingualised forms. According to Bragg et al. \cite{paper39} (2019), there exist a significant amount of limitations that are connected with real-time translation, especially in the context of spoken language and the accessibility of media for people with DHH. These limitations can be reduced by applying deep learning methodologies to convert speech to 3D ASL to improve communication. Additionally, using 3D modeling to translate ASL is more beneficial than a mere 2D representation because this technique is more accurate with regard to sign language's complex and dynamic structure. 3D models also allow us to create animations, video games, augmented reality, and 3D printing. In their journal, Ebling et al. \cite{paper48} (2015) highlight that this 3D(3D avatar) modeling approach can ensure accurate hand gestures and facial expressions for accurate and meaningful translation.

\vspace{0.1cm}
New developments in the 2D sign language recognition format are progressing, yet 3D modeling appears to provide a more precise alternative, particularly for a highly spatial language like ASL. According to Ebling et al. \cite{paper48} (2015), a 3D avatar can obtain signals from subtle facial features and unique hand gestures essential for the accuracy of sign language, which a 2D model fails to depict. Chakladar et al. \cite{paper49} (2021) affirm that the combination of 3D models with NLP might change the whole outlook of symbolization of ASL through dynamic hand gestures during the transition from texts or voice. Nevertheless, the modernistic advanced innovations of breakthrough speech recognition systems, like Whisper AI, increase the applicability of this system in various languages and dialects and make it a more robust and inclusive solution. Studies such as those by Saunders et al. \cite{paper50} (2020) point to their usage of transformer-based systems to deliver both sign accuracy and gesture fluidity in real-time.

\vspace{0.1cm}
Sign language encompasses many levels of complexity that represent a continuous disjoint with the computer world, existing in architectures of their own. Because ASL uses hand gestures, facial interpretation, and body movement simultaneously, such complexities remain highly limiting to the usability of most of the existing technology for real-time application and communication interfaces-and they play a direct role in making it difficult for 2-D models of ASL to capture this depth. As pointed out by Konstantinidis et al. \cite{paper51} (2018), deep learning approaches are still operating in the shallow zone of meeting the most basic of human gestures.

\begin{figure}[h]
    \centering
    \includegraphics[width=0.9\textwidth, height=8cm]{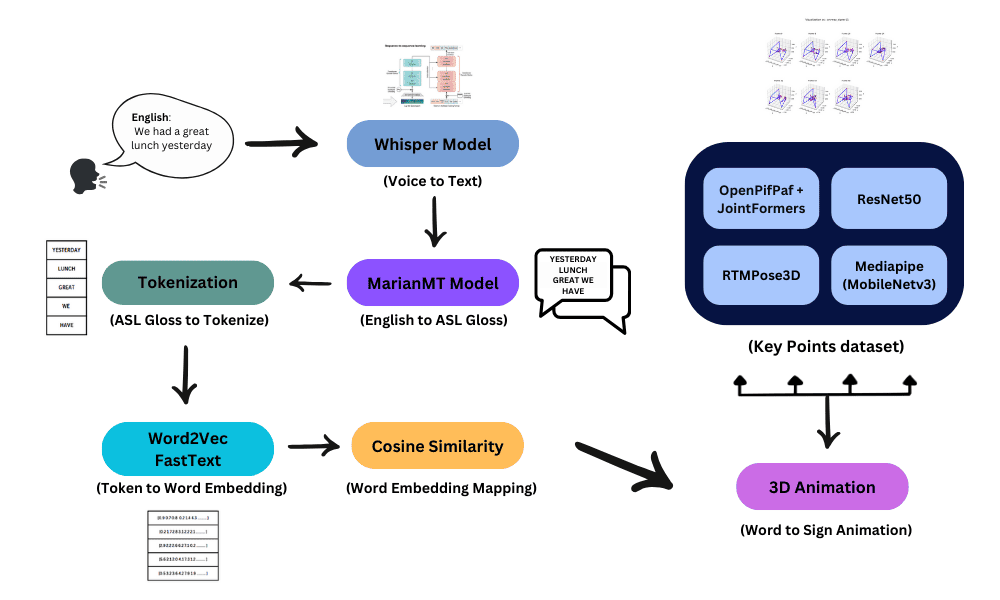}
    \caption{Model Architecture}
    \label{fig:architecture}
\end{figure}

\vspace{0.1cm}
This research aims to develop a comprehensive speech-to-sign language translation system that combines state-of-the-art speech recognition, machine translation, and 3D animation techniques to improve accessibility for the Deaf and Hard of Hearing (DHH) community. The core goal is to evaluate how effectively spoken English can be converted into natural and continuous 3D American Sign Language (ASL) animations. The following objectives guide this work:

\begin{itemize}
\item To build an end-to-end multimodal pipeline that converts English speech into 3D ASL sign language animations using Whisper, MarianMT, and keypoint-based animation.
\item To develop and use a rule-based gloss translation module supported by semantic word embeddings (Word2Vec, FastText) for accurate English-to-ASL gloss generation.
\item To create and utilize two custom datasets: \textbf{BookGlossCorpus-CG} for grammar-aware gloss translation and \textbf{Sign3D-WLASL} for 3D skeletal motion data of ASL signs.
\item To ensure smooth and continuous sign animations through temporal interpolation of skeletal keypoints, producing realistic and natural sign motion.
\item To evaluate the system's performance using metrics such as BLEU score (for gloss translation), animation continuity, and potential for real-time deployment in accessibility tools.
\item To contribute toward inclusive communication technologies by making real-time ASL translation more accurate, natural, and adaptable to real-world use.
\end{itemize}

\section{Related Work}

In 2016, Koller et al. \cite{paper6} (2016) addressed training a CNN for hand images with limited data. To tackle this issue, they integrate a Convolutional Neural Network (CNN) into an iterative Expectation-Maximization (EM) process. The CNN iteratively improves its classification by using weak video-level labels, hence enhancing frame-by-frame annotations with each iteration. Pu et al. \cite{paper7} (2019) propose a new method for continuous sign language recognition (CSLR) by using an iterative alignment network. Their approach addresses CSLR by iteratively improving two components: a 3D-ResNet for learning features and an encoder-decoder with CTC for predicting sequences. A soft-DTW constraint guides training by aligning predicted signs with the ground truth.

\vspace{0.1cm}
Konstantinidis et al. \cite{paper8} (2018) suggest a deep learning technique for recognizing sign language. This method involves analyzing video data, which includes both images and motion, as well as skeletal data, which captures body stance performed using Convolutional Neural Networks (CNN). The system uses separate CNN networks for video and skeletal features, then combines them using various fusion techniques. Yuan et al. \cite{paper9} (2022) propose DeepSign, an advanced deep learning system that utilizes Convolutional Neural Networks (CNNs) to process sign language. DeepSign focuses on two tasks: detecting the presence of sign language in videos and identifying the specific signs being used. This unified CNN-based approach streamlines processing compared to separate models for each task. However, as our method is the opposite of sign language recognition, we want to generate 3D hand signs from voice recognition.

\vspace{0.1cm}
In the late 20th century, many researchers started working with signal processing, specifically converting voice signals to handwritten text. HARPY, \cite{paper10} a significant 1970s speech recognition system, built upon prior systems (Hearsay-1, Dragon) to understand key design choices. Unlike them, HARPY used finite state transition networks for knowledge representation. It employed a beam search strategy to explore promising sound combinations and segmentation for efficiency. The system also incorporated semi-automatic learning of pronunciations and sound representations from training data. Kumar et al. \cite{paper11} (2009) propose an objective method to assess stuttering severity. Traditionally, assessments are subjective, leading to inconsistency. This study tackles this issue by using Mel-Frequency Cepstral Coefficients (MFCCs), a widely used method for extracting features in speech processing, to understand the characteristic's of speech. They likely segment speech into syllables, extract MFCCs, and potentially use machine learning to classify fluent versus stuttered segments. Ravikumar et al. \cite{paper12} (2008) address an objective assessment of stuttering severity in read speech. Traditionally, assessments are subjective. This work targets a common stuttering characteristic - syllable repetitions. They propose the implementation of an automated detection method that leverages Mel-Frequency Cepstral Coefficients (MFCCs) to extract features and employs Dynamic Time Warping (DTW) to compare syllables and identify repeats. This objective approach using MFCCs and DTW offers consistent evaluation compared to subjective methods, potentially improving stuttering assessment.

\vspace{0.1cm}
Mitrovic et al. \cite{paper13} (2009) highlight the significance of feature selection in environmental sound recognition (ESR) because of the diverse range of sounds and the possible constraints of employing generic audio features. The paper highlights the value of statistical analysis (PCA) in identifying redundant features and guiding selection. Chen et al. \cite{paper14} (2021) introduce WavLM, a pre-training approach for various speech-processing tasks. WavLM utilizes self-supervised learning, where a large model is trained on masked speech prediction (predicting missing parts of speech) and denoising (removing noise). Watanabe et al. (2021) \cite{paper15} introduces UniSpeech, a method for learning speech representations that addresses the challenge of limited labeled data. Traditionally, training speech recognition models require a lot of labeled data, which can be expensive. UniSpeech tackles this by proposing a unified approach that leverages labeled and unlabeled data. It uses a multi-task learning framework, combining supervised learning on labeled data (classifying speech sounds) with self-supervised learning on unlabeled data (using contrastive learning). After machine learning, different deep-learning approaches (CNN, LSTM, RNN, etc.) were used for automatic signal processing. Liu et al. \cite{paper16} (2018) address speech recognition in noisy environments. Their approach involves using a Long Short-Term Memory (LSTM) model to improve speech signals before inputting them into a speech recognition system. LSTM networks are adept at handling sequential data, allowing them to identify and remove noise patterns within the speech.

\vspace{0.1cm}
Zhang et al. \cite{paper17} (2019) propose an innovative approach for Automatic Speech Recognition (ASR) by integrating Convolutional Neural Networks (CNNs) with Bidirectional Long Short-Term Memory (LSTM) networks. CNNs are skilled at extracting features from speech signals, while LSTMs excel at capturing context within sequences. This hybrid approach merges these strengths. CNNs likely extract low-level features, and LSTMs then analyze them in both directions to grasp context and long-term dependencies. Li et al. \cite{paper18} (2020) propose ContextNet, a deep learning architecture for ASR that combines CNNs, RNNs, and transducers. While CNNs are powerful for ASR, they can struggle with capturing global context in speech. ContextNet addresses this by incorporating a mechanism using squeeze-and-excitation modules to inject global context information into the CNN layers. The latest approaches are transformer-based approaches. Specific systems are designed to automatically convert spoken or written text into precise representations of a particular sign language, such as American Sign Language (ASL). This emphasizes the ongoing development as a result of generative AI advancements. Camgoz et al. \cite{paper29} (2018) propose an innovative method for Neural Sign Language Translation (NSLT) by using Neural Machine Translation (NMT) approaches. This method directly converts sign language images and videos into spoken language phrases, removing the need for distinct sign recognition. Liu et al. \cite{paper30} (2018) propose an LSTM neural network to enhance speech recognition in noisy environments. This method preprocesses speech signals to remove noise patterns before feeding them into a speech recognition system. It leverages the ability of LSTMs to capture sequential data and distinguish between noise and speech.

\vspace{0.1cm}
Stoll et al. \cite{paper31} (2019) present Text2Sign, a system that converts spoken English into sign language small video clips. Addressing the limitations of data-intensive methods, Text2Sign utilizes a two-stage approach. First, it translates spoken sentences into sign sequences using Neural Machine Translation (NMT). Then, it leverages a Motion Graph and a generative model to translate these signs into realistic videos. Saunders et al. \cite{paper32} (2020) address automatic sign language production limitations. Previous methods focused mainly on hand movements, neglecting facial expressions and mouth motions. This paper proposes Adversarial Training with a transformer-based generator and a discriminator conditioned on the spoken language. The discriminator acts as an "adversary" to assess the realism of generated signs. Chaudhary et al. \cite{paper33} (2023) address a limitation in sign language translation where research often focused only on recognition (sign language to text). The authors present SignNet II, a transformer-based model designed for bidirectional translation. SignNet II can recognize sign language from visuals and generate sign language from text. It achieves this through a two-branched approach: one for sign recognition and another for text-to-sign generation. Wang et al. \cite{paper34} (2022) propose a new method for generating 3D sign language directly from text, addressing the limitations of prior methods that require complex data or architectures. Their approach, called "There and Back Again," utilizes back-translation. Initially, a model is employed to convert text into 2D poses. These poses then train a separate model to convert them back into text. Ultimately, the model undergoes fine-tuning to produce 3D sign language sequences based on the input text.

\section{Methodology}

Our system translates spoken English into 3D animated American Sign Language (ASL). First, speech is converted to text using Whisper. This text is then broken down and converted into an ASL gloss (written representation of signs) using advanced language models and word embeddings. Next, these gloss representations are mapped to 3D key points representing hand and body movements using motion capture data and pose estimation model datasets. Finally, these 3D keypoints generating a real-time animation of the corresponding ASL signs.

\begin{figure}[h]
  \centering
  \includegraphics[width=15cm]{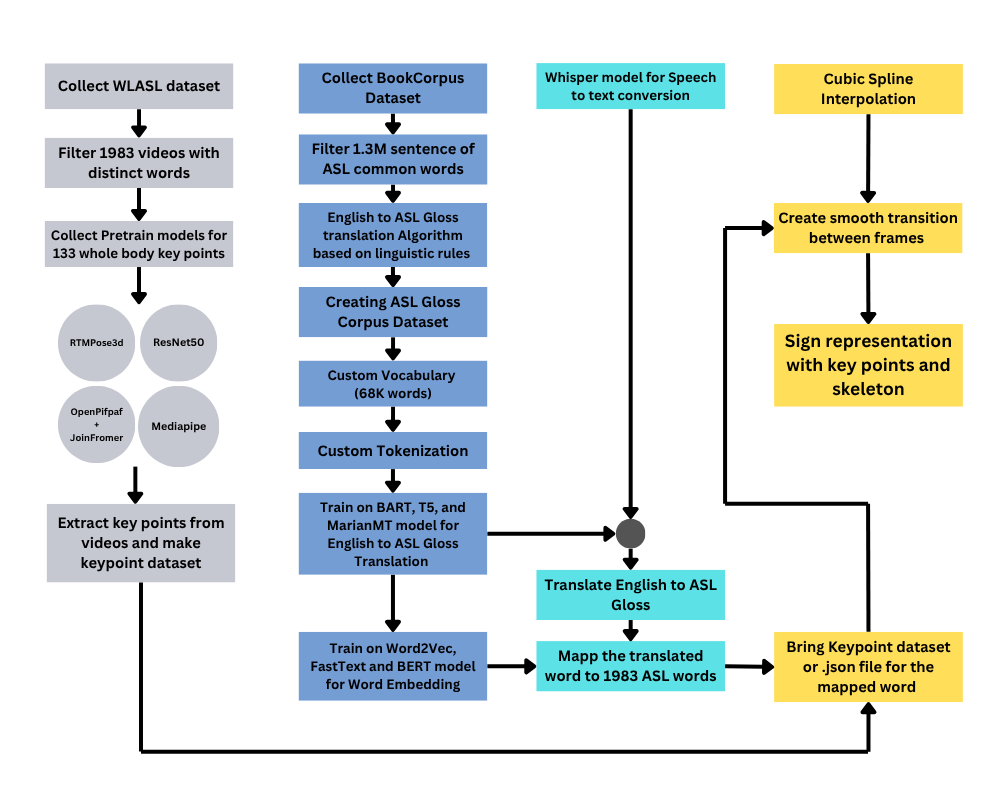}
  \caption{Workflow step by step}
  \label{fig:pipeline}
\end{figure}

\subsection{Voice-to-Text Conversion}
The first step in our pipeline involves converting spoken English input into written text. For this, we utilize Whisper \cite{paper35}, a state-of-the-art automatic speech recognition (ASR) model developed by OpenAI. Whisper is capable of handling diverse accents, background noise, and varying speech patterns, making it well-suited for real-world applications. In this module, the raw audio signal is processed by Whisper to generate a transcript of the spoken sentence. This text output forms the basis for subsequent translation into American Sign Language (ASL) gloss. We perform minimal preprocessing on the audio to preserve its natural characteristics, relying on Whisper’s robust architecture to handle noise and variability. Using Whisper ensures high accuracy and real-time transcription capabilities, which are essential for building an effective and responsive speech-to-sign language system.

\subsection{BookGlossCorpus-CG Dataset}
To facilitate accurate English-to-ASL gloss translation, we introduce BookGlossCorpus-CG, a large-scale parallel corpus derived from the BookCorpus \cite{paper77} dataset. It contains approximately 1.3 million English sentences that have been transformed into ASL gloss using a set of linguistically informed grammar rules. The dataset is grounded in a curated vocabulary of around 2000 high-frequency ASL words, ensuring its relevance for real-world ASL applications.

The transformation process includes several steps designed to align with ASL's grammatical structure. These rules prioritize the temporal marker, topic-comment order, and simplified syntax common in ASL.
\begin{algorithm}[H]
\caption{ASL Gloss Generation for BookGlossCorpus-CG}
\begin{algorithmic}[1]
\For{sentence in BookCorpus}
    \State sentence $\leftarrow$ clean\_text(sentence)
    \State clauses $\leftarrow$ split\_into\_clauses(sentence)
    \For{clause in clauses}
        \State time $\leftarrow$ extract\_temporal\_words(clause)
        \State topic $\leftarrow$ extract\_topic\_nouns(clause)
        \State verb $\leftarrow$ extract\_verbs(clause)
        \State feelings $\leftarrow$ extract\_adjectives\_and\_adverbs(clause)
        \State wh $\leftarrow$ extract\_wh\_questions(clause)
        \State gloss $\leftarrow$ [time, topic, verb, feelings, wh]
        \State gloss $\leftarrow$ apply\_gloss\_formatting(gloss)
        \State save(gloss)
    \EndFor
\EndFor
\end{algorithmic}
\end{algorithm}

By isolating temporal phrases, identifying topical nouns and pronouns, extracting verbs and descriptive terms, and reordering them according to ASL grammar conventions, we produce gloss sentences that closely mimic natural signing.Special tokens are applied to mark fingerspelled lexical items (e.g., \texttt{\#TV}) and proper nouns (e.g., \texttt{fs-EMMA}), which are common in ASL but not directly present in English text. The rule-based logic also accounts for pronoun placement and WH-question positioning—elements that differ significantly between English and ASL. This gloss-converted dataset, BookGlossCorpus-CG, is later used to train our MarianMT translation model. The consistency and scale of the data improve the model’s ability to learn grammar-aware mappings from English to gloss, resulting in high-quality, semantically accurate translations. This step is crucial in enabling our pipeline to support real-time speech-to-sign conversion with clear and fluent sign output.

\begin{figure}[h]
  \centering
  \includegraphics[width=15cm]{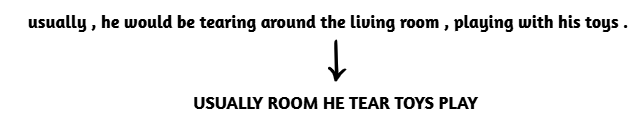}
  \caption{Gloss demo}
  \label{fig:gloss}
\end{figure}

\subsection{Sign3D-WLASL Dataset}
To support realistic 3D gesture generation, we constructed the WLASL \cite{paper42} to Keypoint Dataset, which enriches the original WLASL video dataset with detailed 3D skeletal motion data. The WLASL dataset includes 1,983 carefully curated videos that capture isolated ASL words, signed clearly by native signers. However, the original dataset contains only RGB videos without any annotated body motion data, which limits its direct use for 3D animation. To overcome this, we processed each video frame-by-frame using multiple pose estimation models to extract 133 anatomical keypoints, covering the upper body, facial expressions, and hand articulations. These keypoints are critical for representing the nuances of American Sign Language (ASL), where slight hand and facial movements often change the meaning of a sign. The combination of several pose estimation models ensures robustness across varying lighting, angles, and signer appearances. To improve dataset efficiency and reduce computational overhead, we did not extract keypoints from every single video frame. Instead, we employed a temporal downsampling strategy by selecting frames at regular intervals (e.g., every 4th or 8th frame). This significantly reduced the data volume while retaining the core structure of each sign gesture.

RTMPose3D \cite{paper65} was used as one of the primary methods for extracting full-body 3D keypoints. It operates in two stages: first, RTMDet detects human figures in each frame, and then RTMPose3D predicts 3D keypoints using knowledge learned from the COCO WholeBody dataset. A second keypoint extraction pipeline utilizes a ResNet50-based architecture \cite{paper64} trained on the H3WB dataset, which is designed for detailed upper-body and hand recognition. The ResNet50 backbone extracts spatial features from video frames using deep convolutional layers. OpenPifPaf \cite{paper62} is employed alongside JointFormer \cite{paper63} for bottom-up multi-person keypoint detection. OpenPifPaf detects key body segments and links them into pose structures. It extracts 2D keypoints from each frame, which are then converted to 3D keypoints by JointFormer using spatiotemporal reasoning to refine pose accuracy and completeness. Finally, MediaPipe \cite{paper66} is integrated to support lightweight, real-time keypoint extraction. It uses a multi-stage pipeline that detects human presence, then estimates facial, pose, and hand landmarks.

\subsection{Gloss Translation Model Training}
\subsubsection{Dataset Preparation and Preprocessing}
To enable English-to-ASL gloss translation, we used the BookGlossCorpus-CG dataset, which contains aligned sentence pairs of English input and their corresponding ASL gloss equivalents. Before training, the text was preprocessed through normalization, which included lowercasing, expanding contractions, and removing irrelevant punctuation. Complex sentences were split into simpler clauses to align better with the syntactic structure of ASL, which typically prefers concise, topic-comment-based constructions.

\subsubsection{Tokenization and Input Formatting}
Tokenization was customized for each model architecture. MarianMT \cite{paper71} used a pretrained byte-pair encoding (BPE) vocabulary suited for multilingual translation tasks. In contrast, T5 \cite{paper61} and BART \cite{paper54} utilized SentencePiece tokenizers trained on their respective pretraining corpora. For T5, each English input was prefixed with a task-specific prompt such as "translate English to ASL gloss:", framing the task in a text-to-text manner. All inputs and outputs were padded and truncated to model-specific token limits.

\subsubsection{Model Training and Fine-Tuning}
All models were fine-tuned using standard encoder-decoder architectures with teacher forcing. MarianMT provided strong baseline performance due to its stable translation behavior and multilingual design. T5 generalized well by learning from prompt-based input formatting, while BART produced fluent gloss sequences due to its pretraining on corrupted text reconstruction. Training was monitored using sequence loss, with validation performance checked periodically to avoid overfitting and to ensure stable generalization.

\begin{figure}[h]
    \centering
    \begin{subfigure}[b]{0.3\textwidth}
        \centering
        \includegraphics[width=\textwidth]{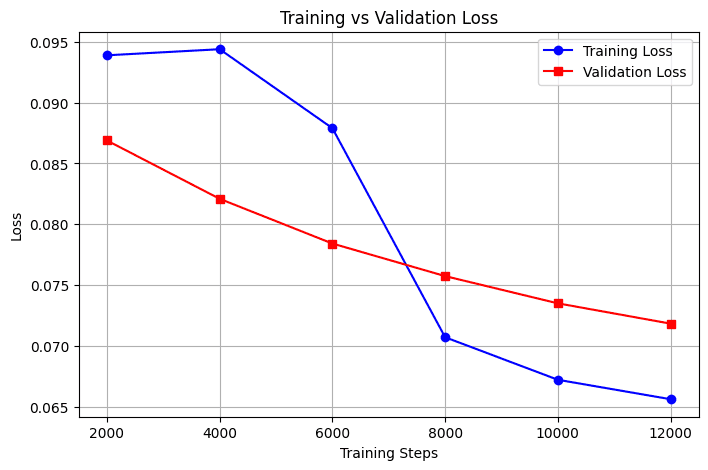}
        \caption{Learning curve of MarianMT}
        \label{fig:marian}
    \end{subfigure}
    \hfill
    \begin{subfigure}[b]{0.3\textwidth}
        \centering
        \includegraphics[width=\textwidth]{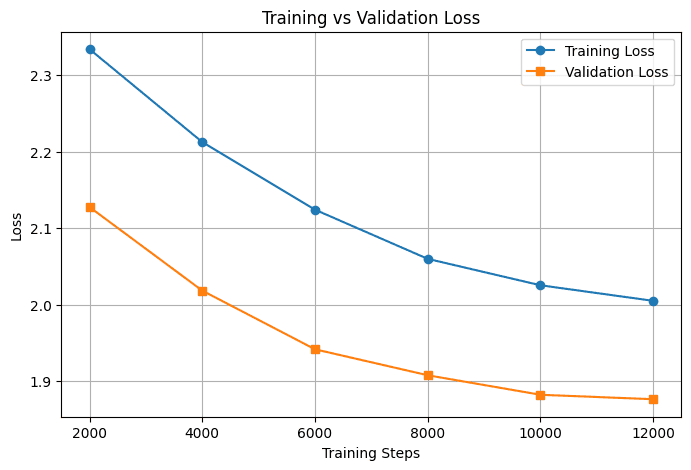}
        \caption{Learning curve of T5}
        \label{fig:t5}
    \end{subfigure}
    \hfill
    \begin{subfigure}[b]{0.3\textwidth}
        \centering
        \includegraphics[width=\textwidth]{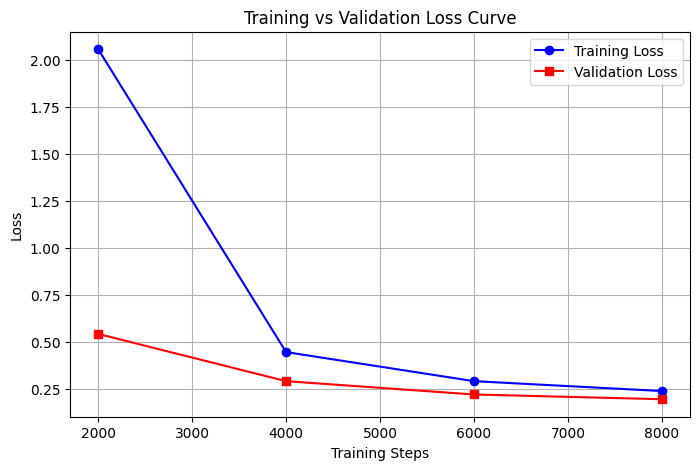}
        \caption{Learning curve of BART}
        \label{fig:bart}
    \end{subfigure}
\end{figure}

\subsubsection{Inference and Postprocessing}
After training, the models generated gloss predictions for new English input. These outputs were decoded from token IDs, cleaned of any special tokens, and postprocessed to match ASL gloss conventions. This included converting certain words to uppercase, marking fingerspelled terms with fs-, and labeling lexical items like \#TV according to ASL standards. These formatted gloss sequences serve as direct input to the 3D animation module.

\subsection{Word Mapping}
To align spoken or written language with ASL vocabulary, we implement a word mapping process that pairs input words with semantically similar ASL gloss terms. Each sentence is tokenized using a pretrained BERT \cite{paper36} tokenizer, and contextual embeddings are generated for each token. These embeddings are compared against a predefined set of common ASL glosses using cosine similarity. The gloss with the highest similarity score is selected for each token, allowing us to normalize different word forms and reduce vocabulary complexity. This results in a cleaner, more consistent gloss sequence that can be effectively used for gesture generation in the 3D animation module.

\subsection{Cubic Spline Interpolation for Smooth Transitions}
To achieve smooth and natural transitions between extracted keypoints in 3D sign language animation, we employ \textbf{cubic spline interpolation}. Unlike simple linear interpolation, which connects two points with a straight line, cubic spline interpolation fits a piecewise series of cubic polynomials between consecutive keypoints. This approach ensures continuity not only in position but also in the first and second derivatives, producing fluid and realistic motion paths. Cubic splines are particularly suitable for interpolating the spatial coordinates of the 3D skeleton joints, as they preserve the natural curvature of motion trajectories while avoiding oscillations or artifacts common in higher-degree polynomial interpolations. By applying cubic spline interpolation, the animation pipeline can generate intermediate frames that fill the gaps caused by frame skipping during keypoint extraction, resulting in a temporally coherent and visually pleasing sign language sequence.

\section{Result}

\subsection{Comparison of BLEU Scores}
BLEU \cite{paper73} is a standard metric for evaluating translation quality by comparing n-gram overlaps between candidate and reference texts. Table~\ref{tab:model_comparison} shows the BLEU scores for T5, BART, and MarianMT models. MarianMT significantly outperforms the others, achieving scores of 0.7714 and 0.8923, indicating better translation accuracy for English to ASL gloss. MarianMT clearly leads in BLEU scores, demonstrating superior performance in English-to-ASL gloss translation. BART performs moderately well but lags behind MarianMT, while T5 scores are significantly lower, likely due to less task-specific optimization.

\begin{table}[h]
    \centering
    \begin{tabular}{lcc}
        \toprule
        \textbf{Model} & \textbf{BLEU-1} & \textbf{BLEU-2} \\
        \midrule
        T5       & 0.0184 & 0.0634 \\
        BART     & 0.5577 & 0.7067 \\
        MarianMT & \textbf{0.7714} & \textbf{0.8923} \\
        \bottomrule
    \end{tabular}
    \caption{BLEU score comparison of translation models.}
    \label{tab:model_comparison}
\end{table}

\subsection{Comparison of MPJPE Values}
Mean Per Joint Position Error (MPJPE) \cite{paper76} measures 3D joint prediction accuracy by calculating the average Euclidean distance between predicted and true joint positions. Lower MPJPE indicates better performance. Table~\ref{tab:mpjpe_comparison} compares an LSTM-based model with a simple Interpolation approach.

\begin{table}[h]
    \centering
    \begin{tabular}{|c|c|}
        \hline
        Model & MPJPE \\
        \hline
        LSTM & 0.044 \\
        Cubic Interpolation & \textbf{0.0429} \\
        \hline
    \end{tabular}
    \caption{MPJPE scores comparison for pose estimation methods.}
    \label{tab:mpjpe_comparison}
\end{table}

Interpolation achieves a slightly better MPJPE (0.0429) than the LSTM (0.044), suggesting it produces more precise joint estimates in this context. For 3D pose estimation, interpolation outperforms the LSTM model slightly in MPJPE, indicating that simpler interpolation can be effective for smooth joint estimation when motions are relatively consistent.

\begin{figure}[h]
    \centering
    \begin{subfigure}[b]{0.45\textwidth}
        \centering
        \includegraphics[width=\textwidth]{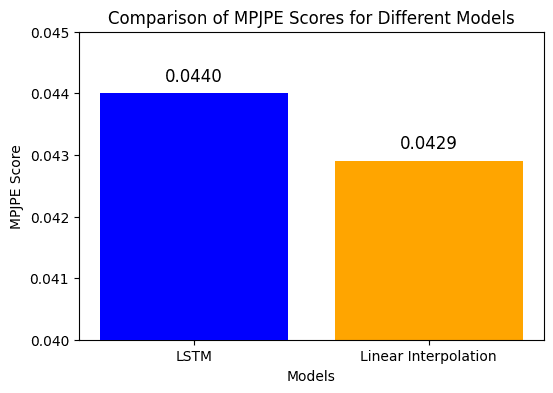}
        \caption{MPJPE comparison between LSTM and Interpolation}
        \label{fig:mpjpe_graph}
    \end{subfigure}
    \hfill
    \begin{subfigure}[b]{0.45\textwidth}
        \centering
        \includegraphics[width=\textwidth]{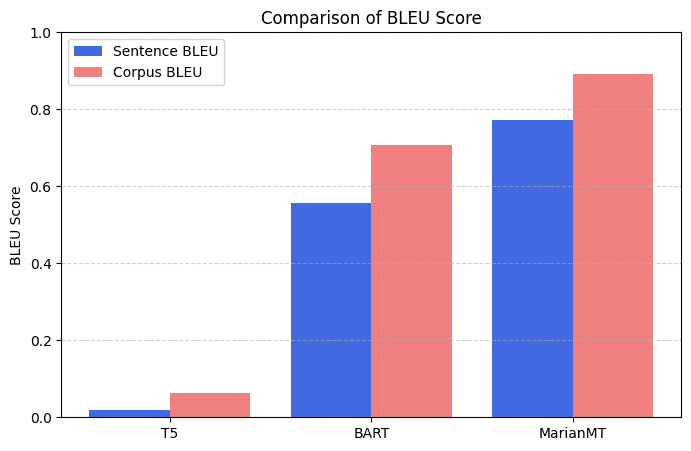}
        \caption{BLEU Scores for different translation models}
        \label{fig:bleu_graph}
    \end{subfigure}
    \caption{Comparison of MPJPE and BLEU scores}
    \label{fig:combined_plots}
\end{figure}

\section{Future Improvements}
\begin{itemize}
    \item Investigate fine-tuning strategies for BART and T5 architectures to enhance their performance on domain-specific English-to-ASL gloss translation tasks.
    \item Develop ensemble methods that synergistically combine the strengths of MarianMT and BART models to improve overall translation robustness and accuracy.
    \item Employ systematic hyperparameter optimization techniques, such as grid search and Bayesian optimization, to identify optimal configurations for translation models.
    \item Advance the English-to-ASL gloss translation framework by incorporating linguistic rules and syntactic structures to more accurately represent ASL grammar and semantic nuances.
    \item Prioritize the development and utilization of domain-adapted models specialized for ASL translation and 3D animation synthesis to achieve higher fidelity and contextual relevance.
\end{itemize}

\section{Conclusion}
This work presents an end-to-end pipeline for translating English speech into 3D American Sign Language animations by integrating state-of-the-art speech recognition, machine translation, word embedding, and 3D keypoint extraction techniques. Through comprehensive dataset creation—including the BookGlossCorpus-CG and Sign3D-WLASL and rigorous model training and evaluation, we demonstrate the effectiveness of transformer-based translation models and interpolation methods in generating accurate and temporally smooth sign language animations. Among the evaluated models, MarianMT exhibits superior translation performance, while cubic interpolation provides precise motion smoothing in 3D pose estimation. Future efforts will focus on refining translation algorithms to better capture ASL grammatical nuances and enhancing model architectures tailored specifically for sign language synthesis. Ultimately, this research contributes to bridging communication gaps and advancing accessible technologies for the Deaf and Hard-of-Hearing community.

\section{Sign3D-WLASL Samples}
\begin{figure}[h]
    \centering
    \begin{subfigure}[b]{0.45\textwidth}
        \centering
        \includegraphics[width=\textwidth]{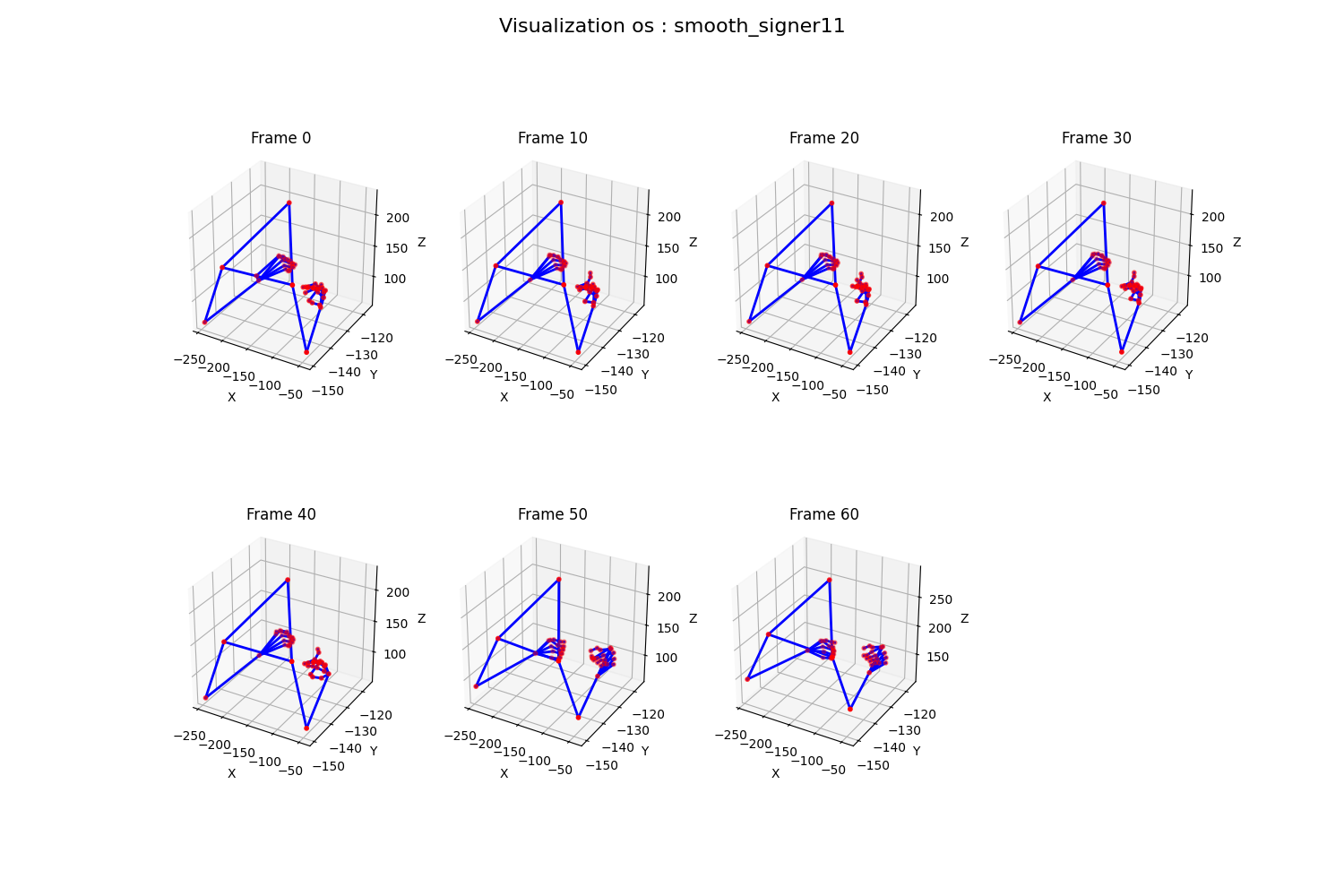}
        \caption{Sample of RMTPose3D}
        \label{fig:rmtpose3d}
    \end{subfigure}
    \hfill
    \begin{subfigure}[b]{0.45\textwidth}
        \centering
        \includegraphics[width=\textwidth]{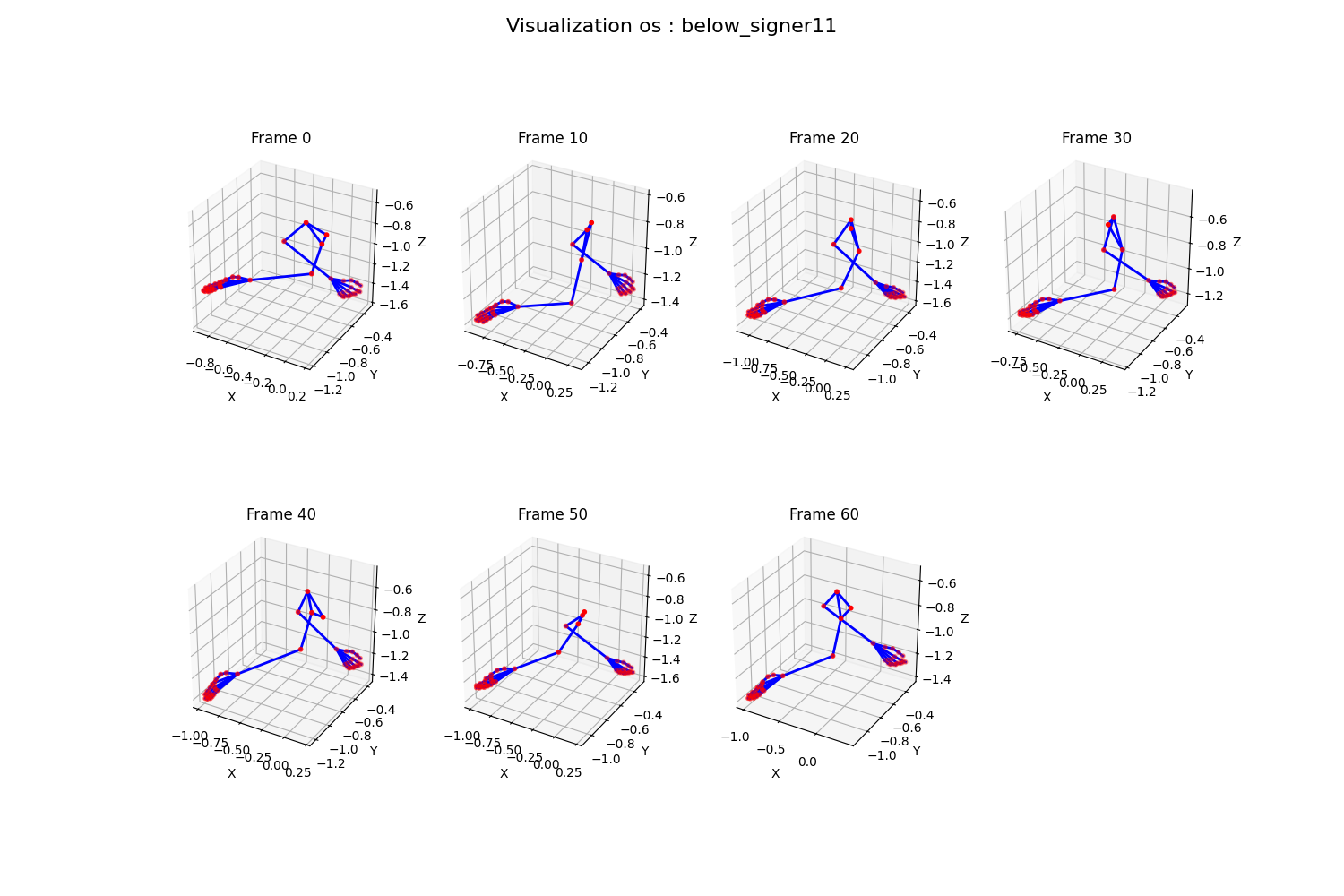}
        \caption{Sample of OpenPifPad and JointFormer}
        \label{fig:openpifpaf}
    \end{subfigure}

    \vspace{0.4cm}

    \begin{subfigure}[b]{0.45\textwidth}
        \centering
        \includegraphics[width=\textwidth]{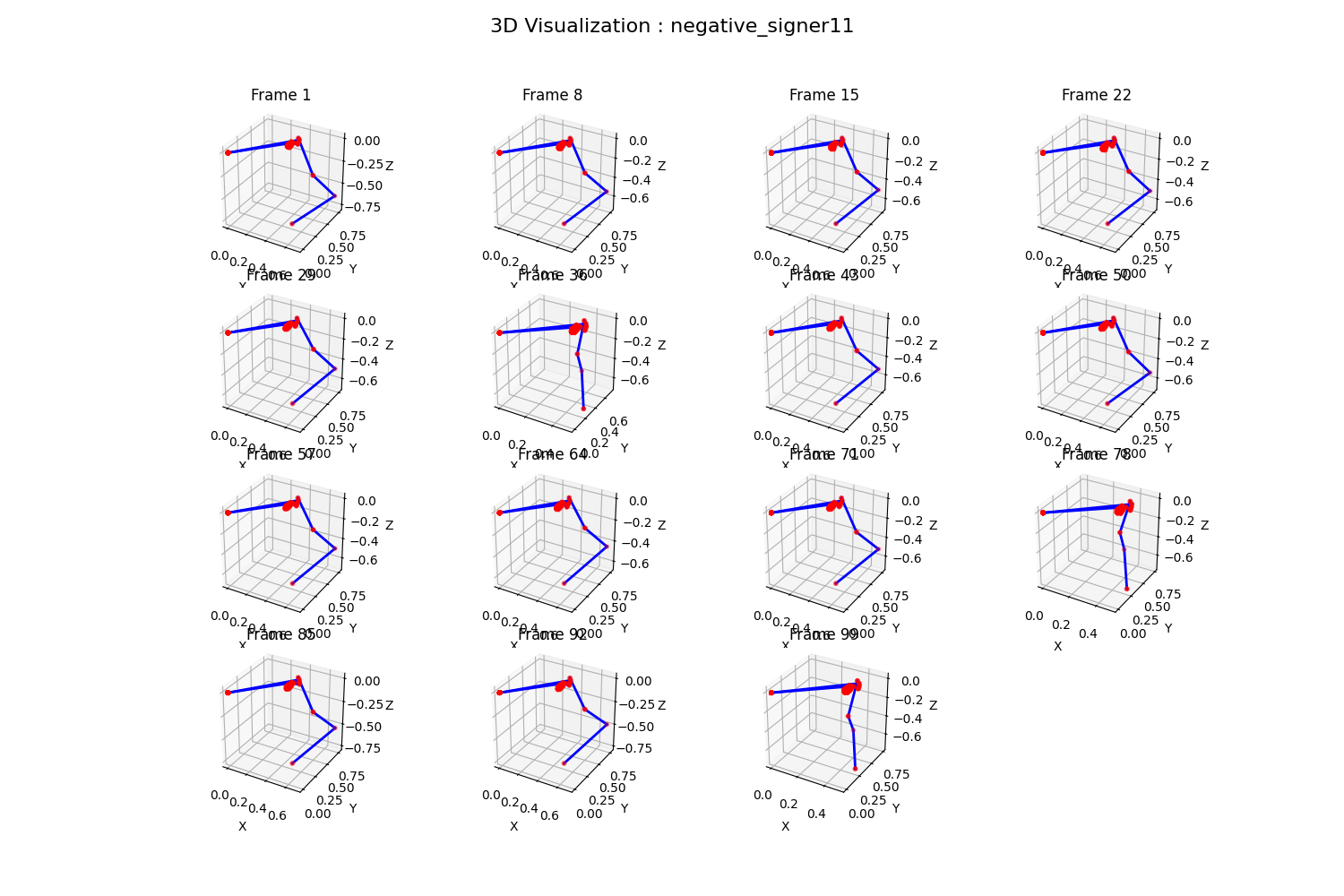}
        \caption{Sample of MediaPipe}
        \label{fig:mediapipe}
    \end{subfigure}
    \hfill
    \begin{subfigure}[b]{0.45\textwidth}
        \centering
        \includegraphics[width=\textwidth]{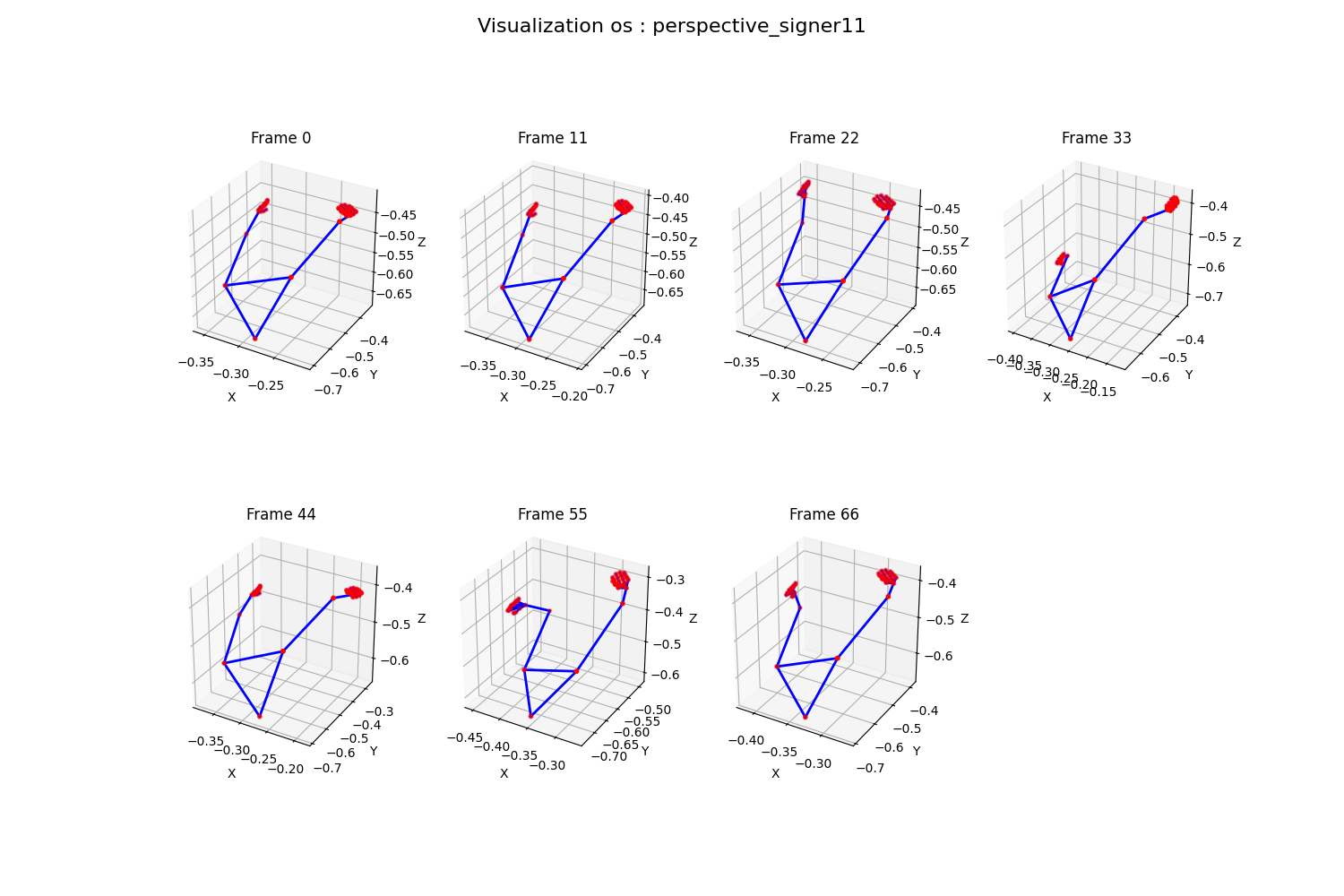}
        \caption{Sample of ResNet-50}
        \label{fig:resnet50}
    \end{subfigure}
\end{figure}

\end{document}